\documentclass[10pt,twocolumn,letterpaper]{article}

\usepackage[pagenumbers]{cvpr} 

\usepackage[dvipsnames]{xcolor}

\definecolor{cvprblue}{rgb}{0.21,0.49,0.74}
\usepackage[pagebackref,breaklinks,colorlinks,citecolor=cvprblue]{hyperref}
\usepackage{multirow}
\usepackage{makecell}
\usepackage{xcolor,colortbl}
\usepackage{color}
\usepackage{comment}
\usepackage[misc]{ifsym}

\definecolor{minetable1colorx}{rgb}{0.75, 0.75, 0.75}

\definecolor{lightgrey}{gray}{0.9}

\title{LMDrive: Closed-Loop End-to-End Driving  with Large Language Models}

\author{%
Hao Shao$^{1, 2}$~~~~~~ Yuxuan Hu$^3$~~~~~~ Letian Wang$^{4}$ \\ \vspace{0.5em}Steven L. Waslander$^{4}$~~ Yu Liu$^{2,5}$~\textsuperscript{\Letter} ~~ Hongsheng Li$^{1,3,5}$~\textsuperscript{\Letter} \\ 
$^1$CUHK MMLab~~~ $^2$SenseTime Research ~~~ $^3$CPII under InnoHK\\$^4$University of Toronto ~~~      $^5$Shanghai Artificial Intelligence Laboratory
}

\begin{document}
\maketitle

\makeatletter{\renewcommand*{\@makefnmark}{}
\footnotetext{\textsuperscript{\Letter} Corresponding author.}\makeatother}

\begin{abstract}
Despite significant recent progress in the field of autonomous driving, modern methods still struggle and can incur serious accidents when encountering long-tail unforeseen events and challenging urban scenarios. On the one hand, large language models (LLM) have shown impressive reasoning capabilities that approach “Artificial General Intelligence”. On the other hand, previous autonomous driving methods tend to rely on limited-format inputs (\textit{e.g.} sensor data and navigation waypoints), restricting the vehicle's ability to understand language information and interact with humans. To this end, this paper introduces LMDrive, a novel language-guided, end-to-end, closed-loop autonomous driving framework. LMDrive uniquely processes and integrates multi-modal sensor data with natural language instructions, enabling interaction with humans and navigation software in realistic instructional settings.  To facilitate further research in language-based closed-loop autonomous driving, we also publicly release the corresponding dataset which includes approximately 64K instruction-following data clips, and the LangAuto benchmark that tests the system's ability to handle complex instructions and challenging driving scenarios.
Extensive closed-loop experiments are conducted to demonstrate LMDrive's effectiveness. To the best of our knowledge, we're the very first work to leverage LLMs for closed-loop end-to-end autonomous driving. Codes can be found at our \href{https://github.com/opendilab/LMDrive}{webpage}.
\end{abstract}
\vspace{-1em}

\section{Introduction}
Remarkable progress in autonomous driving has been witnessed in recent years with an increasing number of commercial autonomous vehicles (AVs) deployed on public roads. Generally, state-of-the-art autonomous driving systems can be categorized into two primary approaches: 1) a modular approach where the system is decomposed into several sub-modules such as perception, prediction, and planning, and fixed interfaces are designed to integrate them together~\cite{li2022bevformer, casas2020implicit}; and 2) an end-to-end approach that directly converts sensor data to control signals via a neural network~\cite{hu2023planning,shao2023reasonnet}. While both of these approaches are widely adopted and constantly making breakthroughs on challenging benchmarks, both of them share a limitation in that they solely rely on fixed-format inputs such as the sensor data, target waypoints, and action commands, which restricts the agent's ability to comprehend multi-modal information and to interact with humans and the environment. On the other hand, large language models (LLMs) have shown an impressive range of capabilities that approach ``Artificial General Intelligence." This encompasses language comprehension, knowledge retrieval, and reasoning. Such capabilities could greatly enhance the safety, controllability, and explanability of autonomous agents. In this work, we seek to answer the question for the first time: \textit{Can we build cognitive autonomous driving systems on top of LLMs, that can interact with human passengers or navigation software simply by natural language?"}

\begin{figure}[t]
  \centering
  \includegraphics[width=1.0\linewidth]{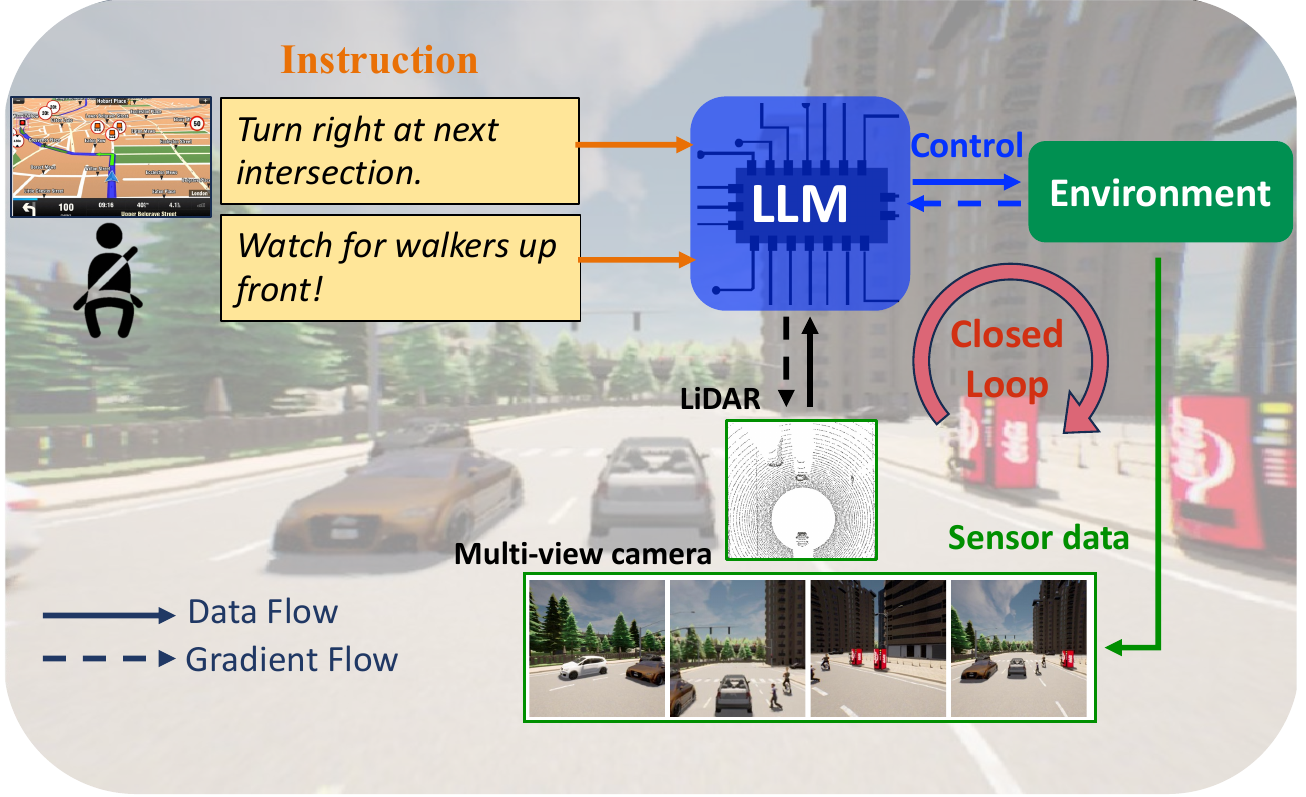}
  \vspace{-1.5em}
  \caption{
We present LMDrive, the first language-guided closed-loop end-to-end driving framework. LMDrive takes as input the language instruction and multi-modal multi-view sensor data, and outputs control signals in real-time to drive in complex scenarios.}
  \label{fig:intro}
  \vspace{-1.5em}
\end{figure}

Making autonomous systems understand natural language opens profound opportunities for advanced reasoning in complex scenarios and efficient interaction with humans, addressing many previously non-trivial problems. 
To name a few: 1) in long-tail unforeseen events and challenging urban situations (\textit{e.g.} complex and dense intersections) where modern AV systems typically struggle~\cite{Jordan2023} or even incur serious accidents~\cite{Trisha2023}, the language-aware AVs can easily survive by following navigation instructions from passengers or navigation software.
2) AVs can adapt to passengers' sudden notice (\textit{e.g.} small objects that are easily missed by perception systems) simply via natural language, which was previously non-trial and required a large amount of hand-crafted rules.

Toward these appealing properties, many pioneering works have explored the potential of using large language models to enhance the AV system's reasoning abilities, interpretability, and overall performance in open-loop settings. One of the most common strategies~\cite{mao2023gpt,chen2023driving,sha2023languagempc,drivelm2023} is to 1) first use LLMs to transform the scene perception results and navigation commands into textual descriptions; 2) feed these textural descriptions into LLMs to generate textual driving decisions; and then 3) transfer textual driving decisions into executable control commands. While good preliminary results are shown, this type of approach, where different LLMs tackle sub-tasks individually, is hard to be trained in an end-to-end manner, loses the capability to scale with a large amount of data, and is not robust to perception errors and uncertainties. For example, since the LLMs in the latter two stages do not have access to the sensor data, inaccurate or missed detections in the first stage can lead to large accumulative errors in the latter stages. 
Towards addressing these issues, end-to-end language-based driving methods~\cite{xu2023drivegpt4} have been proposed.
However, all of these methods undergo training and evaluation in the open-loop setting, where actions are generated and evaluated against the expert actions, but not executed in the actual environments. Notably, when executing navigation instructions such as ``turn right'', the AV agent should not only generate a sequence of actions, but also consider the changes that the actions bring to the environment. The absence of closed-loop evaluation leads to insufficient consideration of critical issues such as cumulative errors, human-robot interaction, and temporal consistency of actions, which makes the resulting methods difficult to scale beyond a short time horizon making them ineffective in actual systems. To the best of our knowledge, there is no existing paper that leverages LLMs for closed-loop end-to-end autonomous driving.

In this work, we introduce LMDrive, an instruction-following multi-modal LLM model for end-to-end closed-loop autonomous driving. The proposed model can process camera-LiDAR sensor data, comprehend driving instructions in natural language, and directly generate vehicle control signals. 
A pre-trained LLM model is adopted and kept frozen to maintain its reasoning capability. To adapt the LLM for autonomous driving, multiple camera-LiDAR data encoders and learnable input/output adapters are integrated. A pre-training strategy specifically designed for the driving task is also introduced for the multi-modal vision encoder.
To facilitate the training of LMDrive in a closed-loop setting, we develop a language-guided driving dataset based on the CARLA simulator~\cite{dosovitskiy2017carla}, which simulates the dynamic world and realistic challenging scenarios. 
To better test the driving model's capability in realistic conversation scenarios, where the instructions might come from humans and navigation software, we 1) consider both navigation instruction and notice instructions; 2) diversify instructions into various phrases via prompt engineering on LLMs; 3) incorporate misleading and unreasonable instructions which are infeasible due to safety concerns or traffic regulations; 4) extend the navigation instructions to include multiple consecutive segments, such as ``Take a left here, then another left at the next one".
Besides, we provide the corresponding LangAuto evaluation benchmark and the pre-trained LMDrive model for reproducibility, and we hope these can facilitate further research in end-to-end closed-loop language-based autonomous driving.

To summarize, this paper makes the following contributions:
\begin{itemize}
    \item We propose a novel end-to-end, closed-loop, language-based autonomous driving framework, LMDrive, which interacts with the dynamic environment via multi-modal multi-view sensor data and natural language instructions. 
    \item We provide a dataset with about 64K data clips, where each clip includes one navigation instruction, several notice instructions, a sequence of multi-modal multi-view sensor data, and control signals. The duration of the clip spans from 2 to 20 seconds.

    \item We present the benchmark LangAuto for evaluating the autonomous agents that take language instructions as navigation inputs, which include misleading/long instructions and challenging adversarial driving scenarios.
    \item We conduct extensive closed-loop experiments to demonstrate the effectiveness of the proposed framework, and analyze different components of LMDrive to shed light on continuing research along this direction. 
\end{itemize}

\section{Related Works}
\subsection{End-to-End Autonomous Driving}
Much progress~\cite{chen2023end, chitta2022transfuser} has been achieved recently in the field of end-to-end autonomous driving.
UniAD~\cite{hu2023planning} devised a framework that incorporates full-stack driving tasks and utilizes query-unified interfaces to communicate between different tasks. 
ThinkTwice~\cite{jia2023think} designed a Look Module to retrieve information from critical regions and utilize the features to refine the coarse prediction.
ReasonNet~\cite{shao2023reasonnet} exploited temporal and global information of the driving scene to improve perception performance and benefit occlusion detection.
InterFuser~\cite{shao2023safety} proposed a transformer-based framework to fully fuse and process information from multi-modal multi-view sensors for comprehensive scene understanding.
TCP~\cite{wu2022trajectory} proposed an approach that integrates the two branches for trajectory planning and direct control by involving a novel multi-step prediction.
LAV~\cite{chen2022learning} introduced some supervisory tasks to learn an viewpoint-invariant representation which can provide a richer supervision signal at training and more information for complex reasoning during inference.
Beyond the previously discussed imitation training methods, several approaches have sought to incorporate reinforcement learning strategies.
Latent DRL~\cite{toromanoff2020end} trained in a supervised way to get a latent representation of the environment observation and conducts reinforcement learning using the representation as the input. 
Roach~\cite{zhang2021end} employed a reinforcement learning agent with privileged access to environmental information and distills a model as the final agent.
ASAPRL~\cite{wang2023efficient} and TaEcRL~\cite{zhou2022accelerating} exploited abstracted skills to effectively improve reinforcement learning efficiency and the final performance by facilitating effective exploration and reward signaling. 
However, these end-to-end methods lack the ability to verbally or textually interact with humans (passengers), and usually have low explanatility in the decision-making process.

\subsection{LLMs in Driving Tasks}
Emerging advancements in large language models (LLMs)~\cite{chung2022scaling,touvron2023llama1,touvron2023llama2,Jaeger2023ICCV,jia2023driveadapter} have been witnessed over the last few months. Furthermore, vision large language models (VLLM) further introduce vision encoders and open the doors for LLMs to interpret not only textual data but also images and data in other modalities~\cite{zhu2023minigpt,liu2023visual,chen2023shikra}.
In the field of autonomous driving (AD), recent research has integrated LLMs into the AD system for better explainability and natural interaction with humans. Some studies adopt the visual language model approach, which can handle multi-modal input data and provide the textual description as well as the control signal for the driving scenarios.
For example, DRIVEGPT4~\cite{xu2023drivegpt4} proposed a multimodal LLM framework, which takes a sequence of frames as input, then generates responses to human inquiries and predicts control signals for the next step. However, since the framework lacks an input command, the predicted control can not follow the specific navigation command, which denotes that the framework is hard to deploy in real scenarios. Meanwhile, more researchers focus on transforming the driving situations into textual descriptions as the input for the LLM, for directly interpreting and reasoning about comprehensive driving situations
In this thread of works, 
GPT-Driver~\cite{mao2023gpt} reformulated motion planning as the task of natural language modeling by converting heterogeneous scene input to language tokens.
LanguageMPC~\cite{sha2023languagempc} leveraged a LLM to reason the complex scenarios and output high-level driving decisions. Then the method tunes a parameter matrix to convert the decision into the low-level control signals.
LLM-Driver~\cite{chen2023driving} utilized the numeric vector as the input modality and fused vectorized object-level 2D scene representation to enable the LLM to answer the questions based on the current environment.

However, this line of work only considered the driving problem in the open-loop settings, and ignored questions such as cumulative error, temporal action consistency, and end-to-end trainability, which are critical for bringing the models into actual closed-loop driving tasks. 
To the best of our knowledge, we are the first language-based end-to-end autonomous driving method in the closed-loop setting. Relevant datasets, benchmarks, and trained models are also open-sourced to facilitate further research in the community.

\section{Dataset generation}
\label{sec:data_generation}

\begin{figure}[t!]
  \centering
  \includegraphics[width=0.99\linewidth]{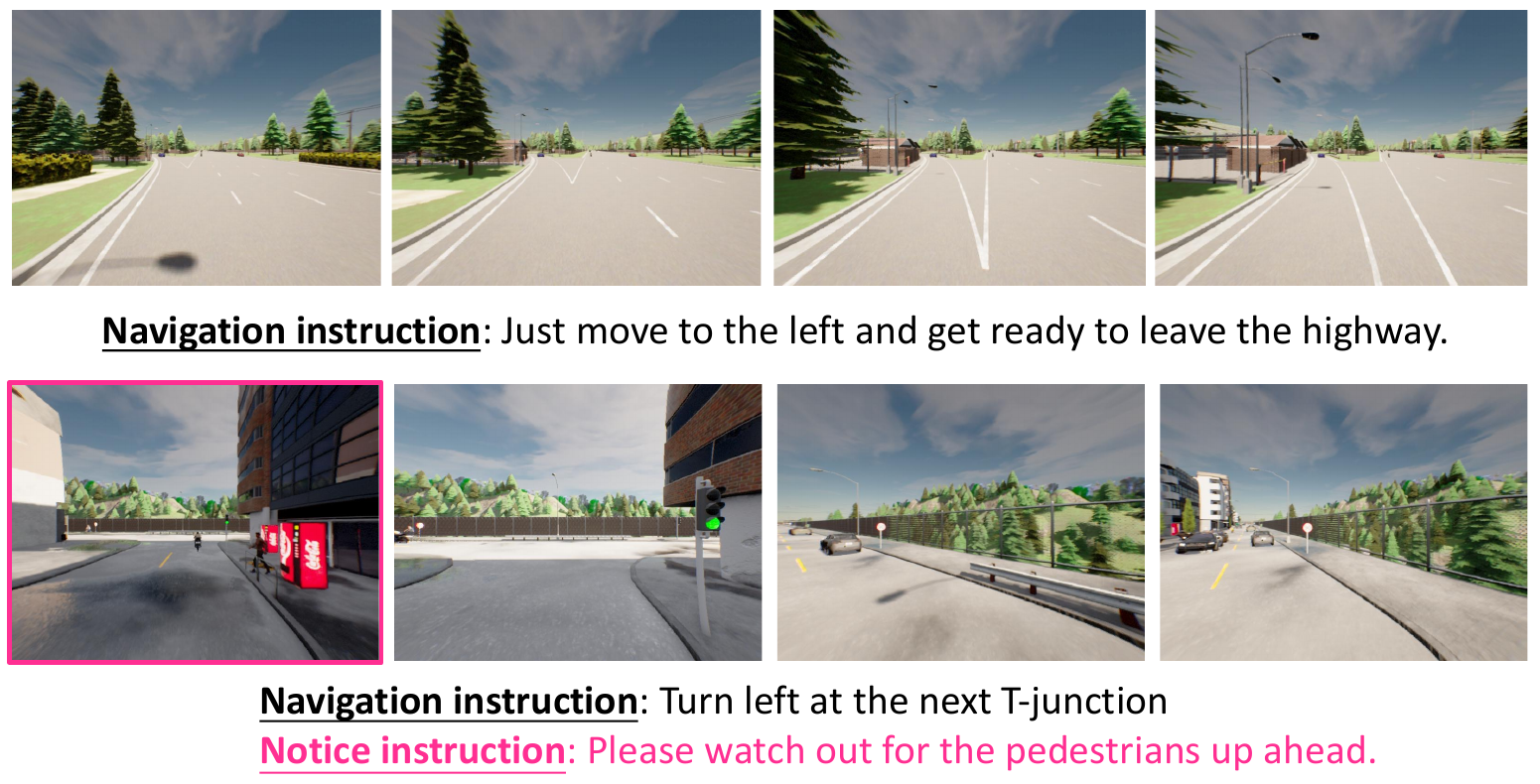}
    \vspace{-0.5em}
  \caption{
  Two examples of the collected data with corresponding labeled navigation instructions and optional notice instructions.
  }
  \label{fig:dataset}
\end{figure}

\begin{figure}[t!]
  \centering
  \includegraphics[width=1\linewidth]{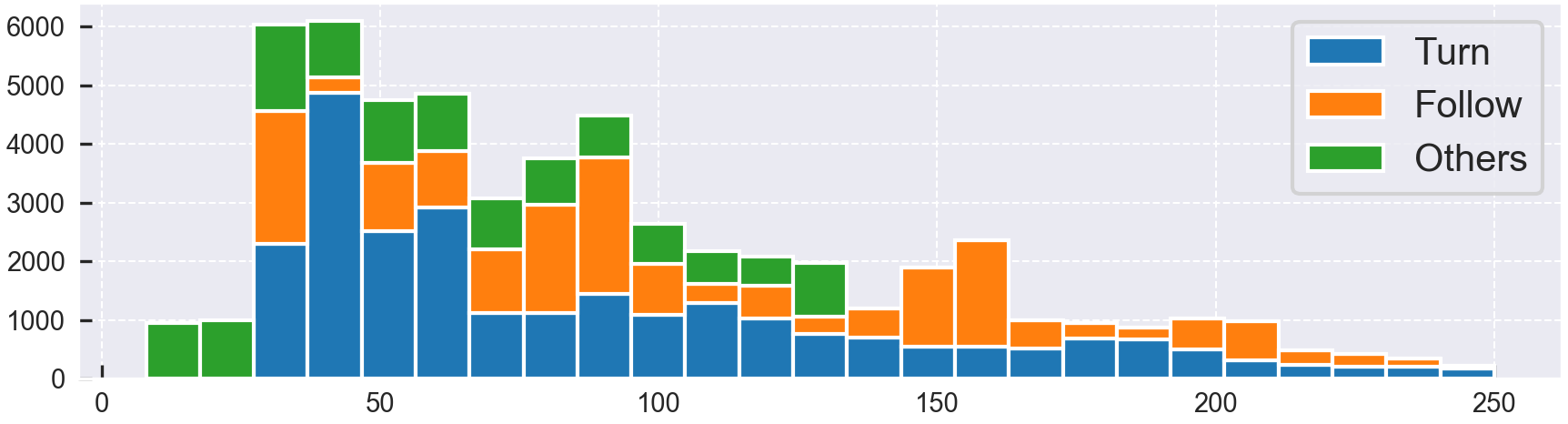}
  \vspace{-2em}
  \caption{
Distribution of parsed clips in terms of clip length and the corresponding navigation instruction type.
  }
  \label{fig:dataset_hist}
  \vspace{-1em}
\end{figure}
We aim to develop an intelligent driving agent that can generate driving actions based on three sources of input: 1) sensor data (multi-view camera and LiDAR), so that the agent can generate actions that are aware of and compliant with the current scene; 2) navigation instructions (\textit{e.g.} lane changing, turning), so that the agent can drive to meet the requirement in natural language (instruction from humans or navigation software); and 3) human notice instruction, so that the agent can interact with humans and adapt to human's suggestions and preferences (\textit{e.g.} pay attention to adversarial events, deal with long-tail events, \textit{etc}).
In this section, we describe how to generate the multi-modal dataset needed to train the agent, and the prompt design for the navigation instruction and human notice instruction. 
Specifically, we choose the CARLA~\cite{dosovitskiy2017carla} as the simulator, because it can simulate a realistic dynamic closed-loop world and it is widely adopted in the field of end-to-end autonomous driving. The data collection consists of two stages: 1) collecting sensor data and control signals with an expert agent; and 2) parsing and labeling collected data with instructions.

\vspace{1mm}
\noindent \textbf{Sensor and control data collection.} We utilize a rule-based expert agent~\cite{shao2023safety} to create a dataset including about 3M driving frames. Since the expert agent can access the privileged information in the CARLA, this dataset will include camera data, LiDAR data, and control actions for each frame. To enhance the diversity of the collected dataset, the agent runs on 2.5k routes, 8 towns, and 21 kinds of environmental conditions (\textit{e.g.} weather, time of the day). We use four RGB cameras (left, front, right, rear) and one LiDAR. The side cameras are angled at 60$^{\circ}$. Besides, we center-crop the front image as an additional focus-view image to capture the status of the distant traffic light. The LiDAR has 64 channels and generates 600K points per second.

\vspace{1mm}
\noindent \textbf{Parsing and language annotation.} In the second stage, we parse the collected data into clips, and label each clip with proper navigation instructions and optional notice instructions. The parsing process takes a sequence of frames as input, and segments these frames into clips, where each clip corresponds to one navigation instruction. For instance, if the agent started to turn left at frame $T_{0}$ and ended at frame $T_{n}$, we will label ($T_{0}$, $T_{n}$) as a new clip with the instruction ``\textit{Hang a left at the next crossroads}''. Besides, if an adversarial event\footnote{The adversarial events include bad road conditions, the front vehicle's sudden brake, unexpected entities rushing into the road from occluded regions, vehicles running a red traffic light, \textit{etc.}} occurs at time $T_{a}$, we will add one notice instruction into this clip, simulating a real-life scenario where a passenger or a side assistance system would communicate with the driver when an emergency happens. 
As shown in Figure~\ref{fig:dataset}, each clip includes sensor data, control signals, the corresponding navigation instruction, and optional notice instructions. The distribution of the parsed clips in terms of clip length and corresponding instruction is shown in Figure~\ref{fig:dataset_hist}.
In our dataset, we collect 64K parsed clips and 464K notice instructions.

\begin{table}[t!]
\centering

\resizebox{\linewidth}{!}{%
\begin{tabular}{l|l}
\toprule
\textbf{Type} & 
\textbf{Three randomly chosen instructions of each instruction type}  \\
\cmidrule(lr){1-1}\cmidrule(lr){2-2}
\multirow{3}{*}{\textbf{Follow}}
& Maintain your current course until the upcoming intersection.  \\
& In [x] meters, switch to left lane.    \\
& Ease on to the left and get set to join the highway.  \\
\cmidrule(lr){1-1}\cmidrule(lr){2-2}
\multirow{3}{*}{\textbf{Turn}}
& After [x] meters, take a left.  \\
& At the next intersection, just keep heading straight, no turn.    \\
& You'll be turning left at the next T-junction, alright?  \\
\cmidrule(lr){1-1}\cmidrule(lr){2-2}
\multirow{3}{*}{\textbf{Others}}
& Feel free to start driving. \\
& Slow down now.    \\
& Head to the point, next one's [x] meters ahead, [y] meters left/right.  \\
\cmidrule(lr){1-1}\cmidrule(lr){2-2}
\multirow{3}{*}{\textbf{Notice}}
& Watch for walkers up front.  \\
& Just a heads up, there's a bike ahead.    \\
& Please be aware of the red traffic signal directly in front of you. \\
\bottomrule
\end{tabular}%
}
\vspace{-2mm}
\caption{
Examples of considered navigation instructions (follow, turn, others) and notice instructions. $\left[ x \right]$ and $\left[ y \right]$ represent the float number for a specific distance.
}
\label{tab:naviation_instruction}
\vspace{-1em}
\end{table}

\vspace{1mm}
\noindent \textbf{Instruction design.}
We consider three types of navigation instructions (follow, turn, and others) along with one type of notice instruction, consisting of a total of 56 different instructions. Table~\ref{tab:naviation_instruction} shows some examples and the full list can be found in supplementary material. 
To enable the agent to drive in realistic instructional settings where the instructions come from navigation software or humans, we 
\begin{itemize}
    \item \textit{Diversifying the instructions}: Considering the inherent richness of natural language, for each type of instruction, we utilized ChatGPT API to generate eight different variants, each carrying the same semantic meaning but varying in phrasing. This enables more comprehensive coverage and flexibility in language interpretation, accommodating the diverse ways the same instruction can be conveyed. 
    \item \textit{Incorporating misleading instructions}: in real-world cases, the navigation software or passengers may give misleading instructions to the AV that violate traffic rules or raise safety concerns. For example, on a single-lane road, following an instruction ``\textit{Change to left lane}'' is dangerous. To improve the robustness of our model against misleading instructions, we simulate these scenarios and add them to our dataset. 
    \item \textit{Connecting multiple instructions:} In many cases, the instructions may consist of two to three consecutive instructions, such as ``\textit{Turn right at this intersection, then go straight to the next intersection and turn right again.}''  We also construct some consecutive complex instruction data to simulate real navigation-based driving scenarios.
\end{itemize}

\section{LMDrive methodology}

\begin{figure*}[t!]
  \centering
  \includegraphics[width=0.95\linewidth]{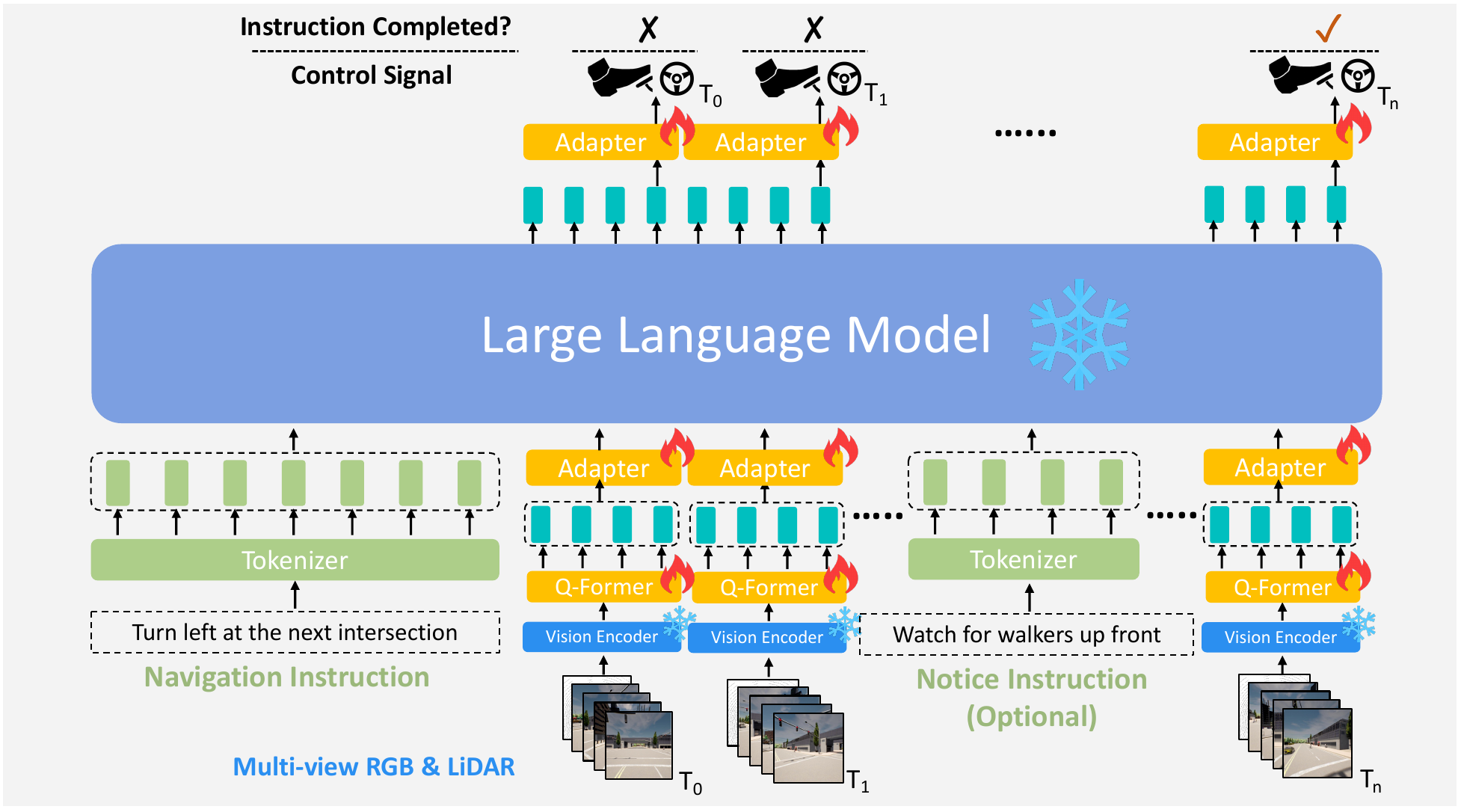}
  \vspace{-2mm}
  \caption{
The structure of the proposed LMDrive model, which consists of two major components: 1) a vision encoder that processes multi-view multi-modal sensor data (camera and LiDAR) for scene understanding and generating visual tokens; 2) a large language model and its associated component (tokenizer, Q-Former, and adapters) that processes all the historic visual tokens and the language instructions (navigation instruction and optional notice instruction), to predict the control signal and whether the given instruction is completed. 
  }
  \label{fig:framework}
  \vspace{-1em}
\end{figure*}

In this work, we propose LMDrive, a framework that can understand and follow high-level driving instructions through natural language.
As illustrated in Figure~\ref{fig:framework}, LMDrive is composed of two major components: 1) a vision encoder that processes multi-view multi-modal sensor data (camera and LiDAR) for scene understanding and generating visual tokens; 2) a large language model and its associated component (tokenizer, Q-Former, and adapters) that takes in the visual tokens and language instruction, to predict the control signal and whether the given instruction is completed. 
We will introduce the vision encoder in Section~\ref{sec:visual_encoder}, and the language model with its associated components in Section~\ref{sec:llm}.
Finally, we describe the training details in Section~\ref{sec:training}.

\subsection{Vision encoder}
\label{sec:visual_encoder}

\begin{figure}[t!]
  \centering
  \includegraphics[width=1.0\linewidth]{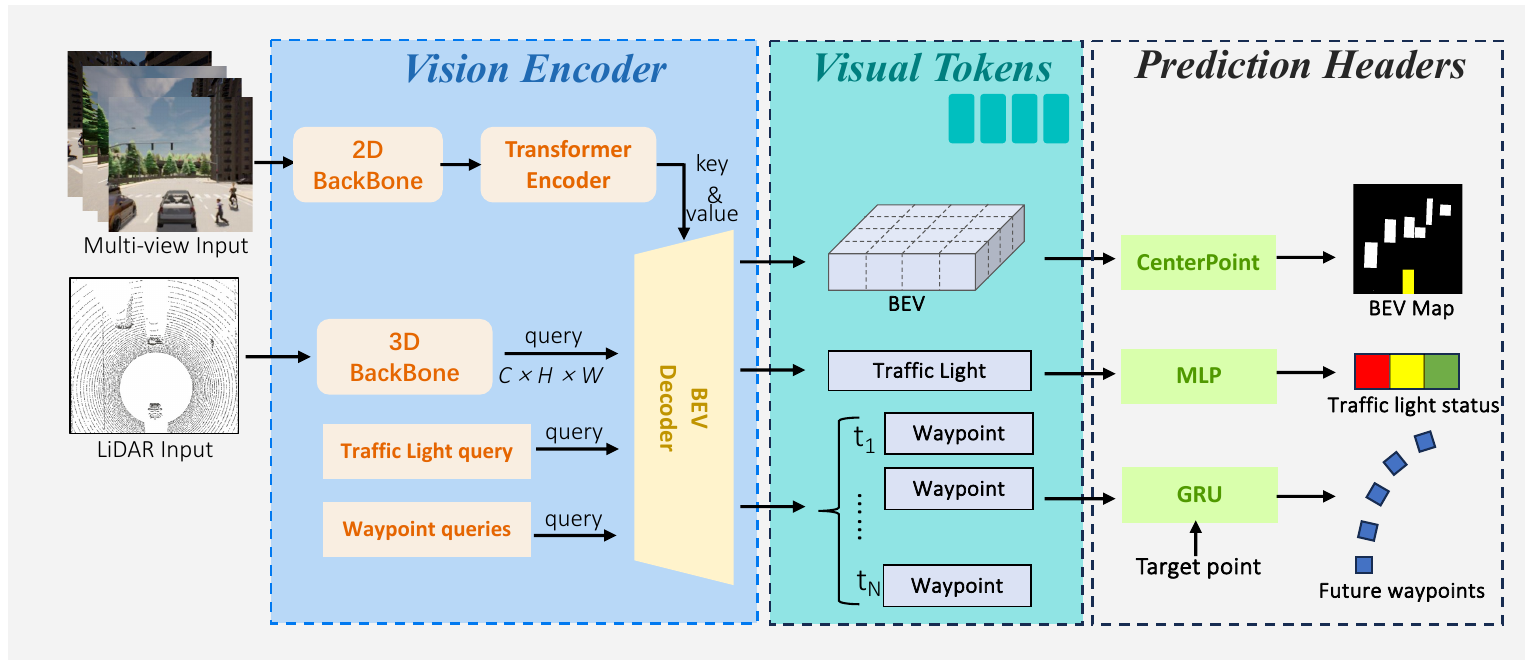}
  \vspace{-2em}
  \caption{
The detailed structure of the vision encoder, which takes as input the multi-view multi-modality sensor data. In the pre-training stage, the vision encoder is appended with prediction headers to perform pre-training tasks (object detection, traffic light status classification, and future waypoint prediction). In the instruction-finetuning stage and inference stage, the prediction headers are discarded, and the vision encoder is frozen to generate visual tokens to feed into the LLM.
  }
  \label{fig:vision_encoder}
  \vspace{-1.5em}
\end{figure}

\vspace{-0.5em}
In the visual language community~\cite{liu2023visual,zhu2023minigpt,chen2023shikra}, the most common way to align vision and language could be using pre-trained CLIP models~\cite{radford2021learning} to encode image features. However, the large flops and parameter size of the CLIP models increase the difficulty of its deployment in AV systems. Also, AV perception systems are usually in 3D to include LiDAR input. 
Hence, inspired by InterFuser~\cite{shao2023safety} and TF++~\cite{jaeger2023hidden}, we design a multi-view multi-modality vision encoder to encode/fuse the sensor data.
As shown in Figure~\ref{fig:vision_encoder}, the vision encoder consists of the sensor encoding part which encodes image and LiDAR input respectively, and a BEV decoder that fuses image and point cloud features to generate visual tokens which are then passed to the language model. Notably, the vision encoder is pre-trained on perception tasks by adding additional prediction heads, then the encoder is frozen for later use by the large language model.

\vspace{1mm}
\noindent \textbf{Sensor encoding.} For each image input, we apply a 2D backbone ResNet~\cite{he2016deep} to extract the image feature map. The feature map is then flattened into one-dimensional tokens. For a comprehensive understanding of the global context from multiple viewpoints, tokens from different views will be fused by a standard $K_{enc}$-layer transformer encoder, each layer containing Multi-Headed Self-Attention~\cite{vaswani2017attention}, MLP blocks and layer normalization~\cite{ba2016layer}.
For the LiDAR input, we adopt a 3D backbone PointPillars~\cite{lang2019pointpillars} to process the raw point cloud data into ego-centered LiDAR features, where each pillar encompasses points within a 0.25m $\times$ 0.25m area. PointNet~\cite{qi2018frustum} is then used to aggregate features and downsample the feature map to $C \times H \times W$, which subsequently serve as BEV queries.

\vspace{1mm}
\noindent \textbf{BEV decoder.} 
The encoded sensor features above are then passed into the BEV decoder to generate visual tokens. 
Specifically, the BEV decoder is designed as a standard transformer with $K_{dec}$ layers.
The BEV point cloud features are fed into the BEV decoder as $H \times W$ queries to attend to the multi-view image features and generate BEV tokens. We also feed $N$ learnable queries and 1 learnable query into the BEV decoder to generate $N$ waypoint point tokens and 1 traffic light token respectively. Thus the three types of visual tokens (BEV, waypoint, and traffic light) will contain rich scene information and will be then presented to the large language model.

\vspace{1mm}

\noindent \textbf{Pre-training with prediction headers.}
we consider three vision encoder pre-training tasks: object detection, future waypoint prediction, and traffic light status classification. 
For object detection, the BEV tokens will pass through a one-stage CenterPoint~\cite{yin2021center} to predict the bounding boxes and velocity of the objects in an $H$m $\times$ $W$m area.
For waypoint prediction, we pass the $N$ waypoint tokens along with the navigation waypoint into the GRU network ~\cite{cho2014learning} sequentially to predict $N$ future waypoint.
For the traffic light status classification, a 2-layer MLP is applied to the traffic light token. 
Three corresponding loss terms are considered: 1) the detection loss as in InterFuser~\cite{shao2023reasonnet}; 2) the $l_1$ waypoint loss; and 3) the cross-entropy traffic light state loss.
Note that these prediction headers are only used in the pre-training of the vision encoder, and will be discarded in the training of the LLMs and inference of the whole model.

\subsection{LLM for instruction-following auto driving}

\label{sec:llm}
As illustrated in Figure~\ref{fig:framework}, in our framework, the LLM functions as the ``brain" throughout the entire driving procedure, processing sensor tokens generated by the frozen vision encoder for each frame, comprehending natural language instructions, generating the necessary control signal, and predicting whether the given instruction is completed.
Specifically, we choose LLaMA~\cite{touvron2023llama1} as the language backbone, which has been widely adopted in many language~\cite{zheng2023judging,geng2023koala} and vision~\cite{liu2023visual,zhu2023minigpt} instruction-tuning models.
We also have three associated components to bridge LLM with instruction, visual information input, and action prediction: 1) a tokenizer, 2) a Q-Former, 3) two adapters.

\vspace{1mm}
\noindent \textbf{Instruction and visual tokenization.} Given the navigation instruction and optional notice instruction, we apply the LLaMA tokenizer~\cite{touvron2023llama1} to convert the instruction into textual tokens. 
Note that the duration of executing one instruction would span from a few seconds to a few minutes, and our model is deployed in the closed-loop setting.
Thus at each frame, we utilize all historic sensor information (with maximum limit $T_{max}$) to depress cumulative error and improve the temporal consistency of the model.
Specifically, for each frame's multi-view multi-modality sensor input, we utilize the vision encoder pre-trained in the previous section to generate visual tokens ($H \times W$ BEV tokens, $N$ waypoint tokens, and one traffic light token). 
However, the number of visual tokens (\textit{e.g.} 406 tokens for each frame) quickly grows too large for the LLM because usually hundreds of frames are needed to complete one instruction.
To overcome this, we follow BLIP-2~\cite{li2023blip} to use the Q-Former to reduce the number of visual tokens. Specifically, for each frame, we employ $M$ learnable queries to attend to the visual tokens via cross-attention layers, which can reduce each frame's visual token number to $M$.
Subsequently, we use a 2-layer MLP adapter to convert the tokens extracted by the Q-Former to share the same dimension as the language token, which can then be fed into the LLM.

\vspace{1mm}
\noindent \textbf{Action prediction.} After receiving a sequence of instructional and visual tokens, the LLM predicts the action tokens. One another 2-layer MLP adapter is then applied to predict future waypoints, as well as a flag to indicate whether the given instruction has been completed. Note that to enhance the supervision signal, we will also conduct prediction for every historic frame during training, and only the prediction of the latest frame will be executed at inference time.  
To get the final control signal, which includes braking, throttling, and steering, following LBC~\cite{chen2020learning}, we use two PID controllers for latitudinal and longitudinal control to track the heading and velocity of predicted waypoints respectively.

\vspace{1mm}
\noindent \textbf{Training objectives.} When finetuning the LLM and its associated components, we consider two loss terms: 1) the $l_1$ waypoint loss; 2) the classification loss (cross-entropy), which determines if the current frame finishes the given instruction.

\subsection{Training details}
\label{sec:training}
LMDrive's training consists of two stages: 1) the vision encoder pre-training stage; and 2) the instruction-finetuning stage, to align the instruction/vision and control signal. 

\vspace{1mm}
\noindent \textbf{Vision encoder pre-training stage.} The vision encoder takes a single frame's sensor data as input, and we use the dataset collected in Section~\ref{sec:data_generation} for training. 
Specifically, since the instruction annotation process will drop some frames, we use the raw dataset before the instruction annotation for the vision encoder pre-training, which includes data of around 3M frames. 
Only the vision encoder is pre-trained with perception tasks for scene understanding.

\begin{table*}[h!]
\centering
\resizebox{0.9\textwidth}{!}{
\begin{tabular}{lccccccccc}
\toprule
\multirow{2}{*}{LLM Backbone} & \multicolumn{3}{c}{LangAuto} & \multicolumn{3}{c}{LangAuto-Short} & \multicolumn{3}{c}{LangAuto-Tiny} \\ \cmidrule(r){2-4} \cmidrule(r){5-7} \cmidrule(r){8-10}
                        &  DS $\uparrow$  & RC $\uparrow$   &  IS $\uparrow$      &  DS $\uparrow$  & RC $\uparrow$   &  IS $\uparrow$  &  DS $\uparrow$  & RC $\uparrow$   &  IS $\uparrow$   \\ \cmidrule(r){1-1} \cmidrule(r){2-4} \cmidrule(r){5-7} \cmidrule(r){8-10}
                  Random Init.      & 10.7$\pm$3.8   &  16.2$\pm$4.9  &  0.63$\pm$0.04  &  14.2$\pm$4.4    &  20.1$\pm$4.4    &  0.72$\pm$0.04   &  20.1$\pm$4.1    &  24.7$\pm$5.1    &  0.75$\pm$0.03     \\
                  LLaMA~\cite{touvron2023llama1}     & 31.3$\pm$1.5   &  37.1$\pm$1.6  & 0.82$\pm$0.01   & 42.8$\pm$7.2     &  49.1$\pm$8.5    &  \textbf{0.87}$\pm$\textbf{0.03}   &  52.2$\pm$5.3    &  57.8$\pm$8.0    &  \textbf{0.91}$\pm$\textbf{0.05}   \\ 
                  LLaMA2~\cite{touvron2023llama2}      & 32.8$\pm$2.1   &  40.1$\pm$2.2  & 0.81$\pm$0.02   &  44.8$\pm$6.2    &  53.5$\pm$5.5    &  0.84$\pm$0.02   &  56.1$\pm$4.1    &  64.2$\pm$4.7    &  0.87$\pm$0.04    \\ 
                  Vicuna~\cite{zheng2023judging}     &  33.5$\pm$1.9  &  39.3$\pm$1.9  &  0.83$\pm$0.02  &   45.3$\pm$4.9  &  54.3$\pm$3.9    &  0.83$\pm$0.03 &  55.5$\pm$3.9    &  63.1$\pm$4.2    &  0.88$\pm$0.04    \\ 
                  Vicuna-v1.5~\cite{zheng2023judging}     &  34.0$\pm$3.8  &  39.0$\pm$3.3  &  \textbf{0.85}$\pm$\textbf{0.06}  &   47.0$\pm$4.3   &  56.5$\pm$2.4    &  0.83$\pm$0.04    &  59.0$\pm$2.6    &  69.9$\pm$2.3    &  0.84$\pm$0.02    \\ 
                  LLaVA-v1.5~\cite{liu2023improved}      &  \textbf{36.2}$\pm$\textbf{2.3}  &  \textbf{46.5}$\pm$\textbf{4.3}  &  0.81$\pm$0.03  &   \textbf{50.6}$\pm$\textbf{1.7}   &  \textbf{60.0}$\pm$\textbf{3.4}    &  0.84$\pm$0.04   &  \textbf{66.5}$\pm$\textbf{3.6}    &  \textbf{77.9}$\pm$\textbf{2.3}    &  0.85$\pm$0.02    \\ 
                        \bottomrule
\end{tabular}}
\vspace{-2.5mm}
\caption{Performance comparison of 6 LLM backbones on the LangAuto benchmark. We report the metrics for 3 evaluation runs.}
\label{table:llm_backbone}
\vspace{-1em}
\end{table*}

\begin{table}[]
\centering
\resizebox{0.9\linewidth}{!}{
\begin{tabular}{cccc}
\toprule
Module design  &  DS $\uparrow$  & RC $\uparrow$   &  IS $\uparrow$       \\ \cmidrule(r){1-1} \cmidrule(r){2-4} 
                  Baseline (LLaVA-v1.5)       & \textbf{36.2}$\pm$\textbf{2.3}  &  \textbf{46.5}$\pm$\textbf{4.3}  &  \textbf{0.81}$\pm$\textbf{0.03}  \\
                  w/o Q-Former     & 31.7$\pm$3.5   &  41.2$\pm$4.4  & 0.79$\pm$0.02      \\ 
                  w/o using BEV tokens     & 33.9$\pm$3.9   &  45.9$\pm$5.1  & 0.72$\pm$0.03  \\ 
                  w/o visual pre-training     & 16.9$\pm$5.1   &  24.1$\pm$4.7  & 0.70$\pm$0.04  \\ 
                  
                        \bottomrule
\end{tabular}}
\vspace{-2.5mm}
\caption{Ablation study on the module design.}
\vspace{-1em}

\label{table:model_architecture}
\end{table}

\begin{table*}[h!]
\centering
\footnotesize
\resizebox{0.9\linewidth}{!}{
\begin{tabular}{l|cccccccc}
    \toprule
     \thead{LLM Backbone} & \thead{Benchmark Type} & \thead{Infraction \\ Score $\uparrow$} & \thead{Vehicle \\ Collisions $\downarrow$} & \thead{Pedestrian \\ Collisions $\downarrow$} & \thead{Layout \\ Collisions $\downarrow$} & \thead{Red light \\ Violations $\downarrow$} & \thead{Offroad \\ Infractions $\downarrow$} & \thead{Blocked \\ Infractions $\downarrow$} \\
    \cmidrule(r){1-2}
    \cmidrule(r){3-9}
    \multirow{2}{*}{LLaVA-v1.5} & LangAuto  & 0.81 & 0.33 & 0.03  & 0.50 & 0.92 & 0.36 & 0.22 \\
    & \cellcolor{lightgrey}LangAuto-Notice & \cellcolor{lightgrey}0.87 & \cellcolor{lightgrey}0.17 & \cellcolor{lightgrey}0.02  & \cellcolor{lightgrey}0.31 & \cellcolor{lightgrey}0.50 & \cellcolor{lightgrey}0.17 & \cellcolor{lightgrey}0.27	\\ \midrule
    \multirow{2}{*}{Vicuna-v1.5} & LangAuto & 0.85 &  0.30  & 0.03 & 0.43 & 1.18 & 0.24 & 0.19 \\
     & \cellcolor{lightgrey}LangAuto-Notice & \cellcolor{lightgrey}0.91 & \cellcolor{lightgrey}0.15 & \cellcolor{lightgrey}0.01 & \cellcolor{lightgrey}0.28& \cellcolor{lightgrey}0.56 & \cellcolor{lightgrey}0.26 & \cellcolor{lightgrey}0.19 \\
    \bottomrule
\end{tabular}}
\vspace{-2mm}
\caption{Performance comparison on LangAuto and LangAuto-Notice benchmarks. The metrics (except infraction score) are normalized by the driven distance (km).}
\label{table:notice_benchmark_notice}
\vspace{-1em}
\end{table*}

\begin{table}[h!]
\centering
\footnotesize
\resizebox{0.9\linewidth}{!}{
\begin{tabular}{l|cccc}
    \toprule
     LLM Backbone & Benchmark Type & DS $\uparrow$ & RC $\uparrow$ & IS $\uparrow$  \\
    \cmidrule(r){1-2}
    \cmidrule(r){3-5}
    \multirow{2}{*}{LLaVA-v1.5} & LangAuto  & 36.2 & 46.5 & 0.81 \\
    & \cellcolor{lightgrey}LangAuto-Sequential  & \cellcolor{lightgrey}34.0 & \cellcolor{lightgrey}43.7 & \cellcolor{lightgrey}0.81 	\\ \midrule
    \multirow{2}{*}{Vicuna-v1.5} & LangAuto& 34.0 & 39.0 & 0.85  \\
     & \cellcolor{lightgrey}LangAuto-Sequential  & \cellcolor{lightgrey}31.9 & \cellcolor{lightgrey}37.1 & \cellcolor{lightgrey}0.84 \\
    \bottomrule
\end{tabular}}
\vspace{-2.5mm}
\caption{Performance comparison on LangAuto and LangAuto-Sequential benchmarks.}
\label{table:notice_benchmark_sequential}
\vspace{-1em}
\end{table}

\vspace{1mm}
\noindent \textbf{Instruction-finetuning stage.} The entire system is trained for end-to-end autonomous driving under the guidance of the instruction, where the Q-Former and Adapters are trainable and the other components are frozen. While our LMDrive takes a sequence of frames as input, during training we set a fixed sequence length $T_{max}$ for building up batch data. The training utilizes the instruction-following data generated in Section~\ref{sec:data_generation}. 
To enable the model to reject misleading instructions, we label the corresponding data as  `completed' after the misleading instruction is given for about 1 second.
Since the dataset is collected at a high frequency ($\sim$10Hz), the data in adjacent frames are highly similar.
To encourage efficient training, following video prediction methods~\cite{wang2016temporal,feichtenhofer2019slowfast, shao2020temporal}, we sample the training frames in a fixed interval, and apply temporal augmentation which randomly shifts the training frames either forward or backward, with the random shift less than the fixed interval.

\section{LangAuto Benchmark}

We propose the LangAuto (Language-guided Autonomous Driving) CARLA benchmark, the first benchmark that evaluates closed-loop driving performance under language instructions.
Compared with the previous CARLA benchmark, Town05~\cite{Prakash2021CVPR} and Longest6~\cite{chitta2022transfuser} that navigate the agent with discrete driving commands or target waypoints, our benchmark only provides the AV with a navigation instruction and optional notice instructions in natural language.

Specifically, the LangAuto benchmark covers all 8 publicly available towns in CARLA to include various scenarios (\textit{e.g.} highways, intersections, roundabouts). 
We also consider 16 kinds of environmental conditions, encompassing the combinations of 7 weather conditions (Clear, Cloudy, Wet, MidRain, WetCloudy, HardRain, SoftRain) and 3 daylight conditions (Night, Noon, Sunset).
Besides, LangAuto consists of three tracks to fully test the agent's instruction-following abilities:
\begin{itemize}
    \item LangAuto track: For each route, navigation instructions are given and updated to the agent based on the agent's current position. We also divide this track into three sub-tracks with different route lengths, to better distinguish the performance. LangAuto where the routes are longer than 500 meters, LangAuto-Short where the route length is between 150 and 500 meters, and LangAuto-Tiny where the route length is shorter than 150 meters.
    \item LangAuto-Notice track: Based on the LangAuto track, we additional add notice instructions to the agent. This setting simulates real cases where passengers or other assistance systems can give real-time notice in long-trail complex or adversarial scenarios, which is usually hard for the AV system to handle by itself.
    Ideally, the agent that can comprehend and leverage instruction can achieve better performance.
    \item LangAuto-Sequential track: Based on the LangAuto track, we merge 10\% of consecutive 2 to 3 instructions into a single long instruction. This setting mimics the realistic scenarios where the multi-sentence instructions come from the passengers or navigation software.
\end{itemize}
Note that misleading instructions will be randomly ($\sim$ 5\%) and intermittently given to the driving agent, which lasts for a certain duration (1-2 seconds). The driving agent is expected to reject these misleading instructions and execute safe actions that are compliant to the current scene, until the next correct instruction is spawned.

\vspace{1mm}

\noindent  \textbf{Metrics.} We consider three major metrics introduced by the CARLA LeaderBoard~\cite{leaderboard}: route completion (RC), infraction score (IS), and driving score (DS). The route completion refers to the percentage of the total route length that has been completed.
It only takes into account the distance traveled along the predetermined route, where each segment of the predetermined route corresponds to a navigation instruction.
If the agent deviates too far from the route, the agent is regarded as violating the instruction, and this episode is marked as a failure and terminated.
The infraction score measures infractions triggered by the agent. When collisions or traffic rule violations occur, the infraction score is decayed by a corresponding discount factor.
The driving score is the product of the route completion ratio and the infraction score, describing both driving progress and safety. It is generally recognized as the primary ranking metric.

\section{Experiments}
\subsection{Experiment Setup}
We implement and evaluate our approach on the open-source CARLA simulator of version 0.9.10.1~\cite{dosovitskiy2017carla}. 
For the 2D backbone and 3D backbone of the vision encoder, ResNet-50~\cite{he2016deep} backbone is pre-trained on ImageNet, and PointPillars~\cite{lang2019pointpillars} is trained from scratch. 
$C, H, W, N$ are set as 256, 50, 50, 5 respectively. 
For the Q-former token number $M$, we empirically found that $M=4$ tokens achieve decent performance. 
During the instruction-finetuning stage, we sample the training frames in a fixed interval of 2. 
Because each parsed clip typically contains multiple notice instructions, to avoid overfitting to notice following and generating over-conservative behaviors, during training we randomly removed notices for 75\% of the clips. For the remaining clips, we ensure that a maximum of one notice is included. We refer readers to the appendix for more details.

\subsection{Quantitative Results}

\noindent \textbf{LLM backbones.} In Table~\ref{table:llm_backbone}, we evaluate our method with different pretrained or randomly initialized LLM models of 7B parameters:
LLaMA~\cite{touvron2023llama1} and LLaMA2~\cite{touvron2023llama2} were pretrained with large public language datasets; Vicuna~\cite{zheng2023judging} and 
Vicuna-v1.5~\cite{zheng2023judging} are additionally finetuned with human conversation data; LLaVA-v1.5~\cite{liu2023improved} incorporate multi-modal data (\textit{e.g.} language, images) for training. We observe that LLaVA-v1.5~\cite{liu2023improved} surpasses the other LLM models, which demonstrates the importance of adopting pretrained multi-modal LLMs for instruction-following autonomous driving. Besides, the models finetuned with instruction data perform better than others (Vicuna-v1.5 $>$ LLaMA2 $\approx$ Vicuna $>$ LLaMA). 
We also test a randomly initialized 7B LLM model, which struggles to drive properly with the same amount of training data, which demonstrates the necessity of finetuning pretrained LLM for instruction-following driving.

\vspace{1mm}
\noindent\textbf{Module Design.} In Table~\ref{table:model_architecture}, we conduct ablation studies on different components. First, instead of reducing the number of BEV tokens with the Q-Former, we directly downsample the BEV features to 4 $\times$ 4, then feed them into the LLM  (denoted as ``w/o Q-Former''). The average driving score dropped from 36.2 to 31.7. 
Second, excluding the BEV tokens input into LLM decoder (denoted as ``w/o using BEV tokens'' results in a decreased infraction score (0.81 to 0.72). One explanation is that the BEV tokens are important for detecting and reasoning the surrounding obstacles and road structures.
Third, we remove the pre-training stage of the vision encoder and directly train it from scratch in the instruction-finetuning stage. The driving score drops to 16.9 (denoted as ``w/o visual pre-training''), demonstrating the importance of our proposed vision encoder pre-training.

\vspace{1mm}
\noindent\textbf{LangAuto-Notice Benchmark.} The LangAuto-Notice benchmark provides some notice instructions to the agent when adversarial events happen. As shown in Table~\ref{table:notice_benchmark_notice}, our agents can effectively leverage the information of the notice in real-time, resulting in a significant decrease in both collisions and traffic rule violations.

\vspace{1mm}
\noindent\textbf{LangAuto-Sequential Benchmark.} In Table~\ref{table:notice_benchmark_sequential}, we demonstrate the effect of the LangAuto-Sequential benchmark where some consecutive 2 to 3 navigation instructions are merged together to a long and complex instruction. 
This setting additionally requires the agent to be able to be temporally aware of which instruction has been completed and which has not. 
Our agents based on LLaVA and Vicuna both have a drop in the driving score and route completion ratio.

\begin{figure}
  \centering
  \includegraphics[width=\linewidth]{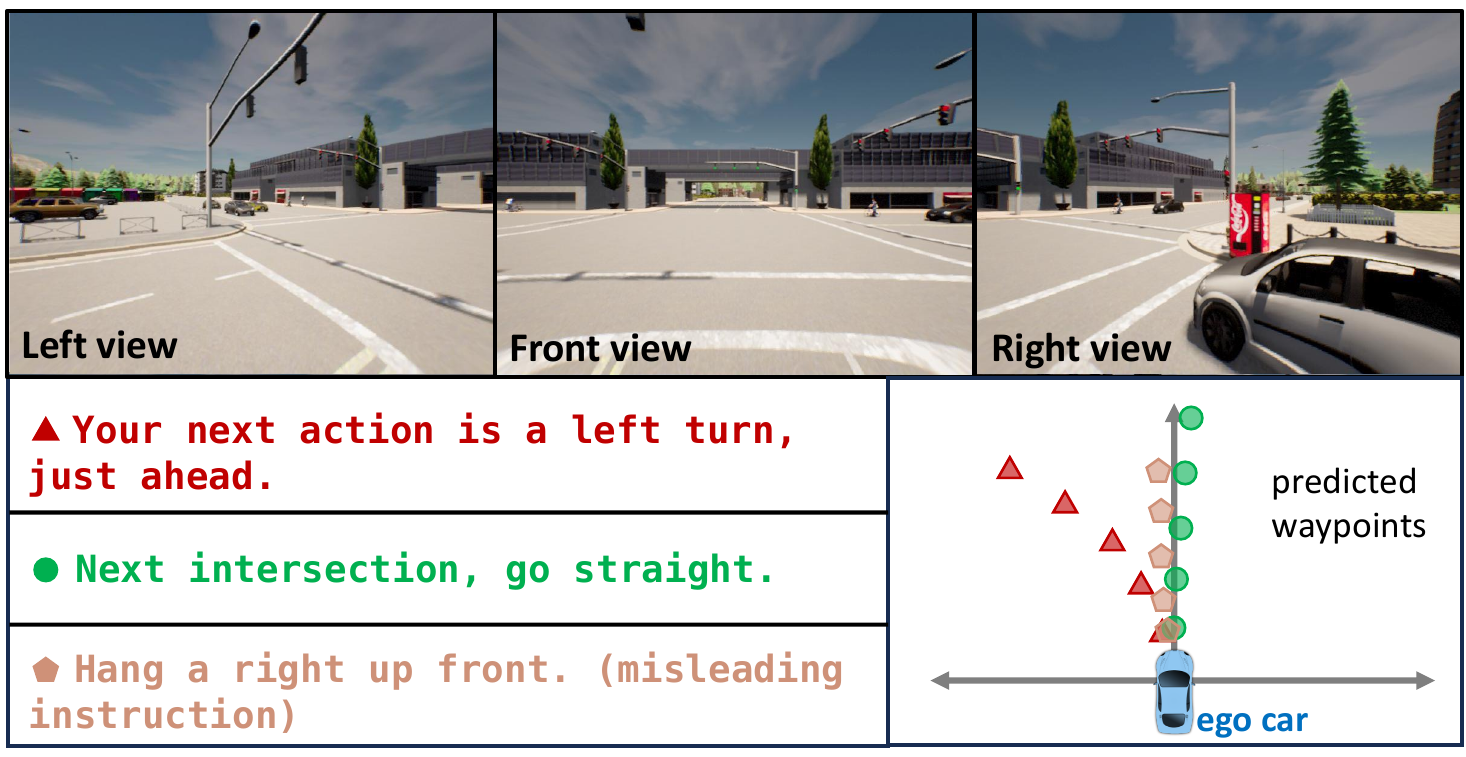}
  \vspace{-2em}
  \caption{An example of how our LMDrive predicts future waypoints, given sensor inputs and varied navigational instructions. Under the first two instructions, our model predicts different waypoints accordingly. The third instruction is a misleading one (turn right on a left-only lane). The model appropriately rejects the incorrect instruction, generating a slower speed and a safe path that is compliant with the scenario.
  }
  \label{fig:vis_different_instruction}
  \vspace{-1em}
\end{figure}

\section{Conclusion}
In this paper, we introduced LMDrive, a language-guided, end-to-end, closed-loop autonomous driving framework. LMDrive incorporates natural language instructions along with multi-modal sensor data, enabling human-like interaction and navigation in complex driving scenarios. We also propose the language-guided driving dataset, comprising around 64K multi-modal data clips along with corresponding navigation instructions. We established the LangAuto benchmark for evaluating autonomous driving systems considering natural language instructions. The effectiveness of LMDrive was demonstrated through extensive closed-loop experiments, underlining the potential of improving the interaction of autonomous vehicles with humans and the environment. Our work serves as an encouraging starting point for further exploration and developments in the field of language-based closed-loop end-to-end autonomous driving.

{
    \small
    \bibliographystyle{ieeenat_fullname}
    \bibliography{main}
}

\clearpage
\appendix
\setcounter{page}{1}
\maketitlesupplementary

\section{Implementation Details}
\begin{figure}
  \centering
  \includegraphics[width=\linewidth]{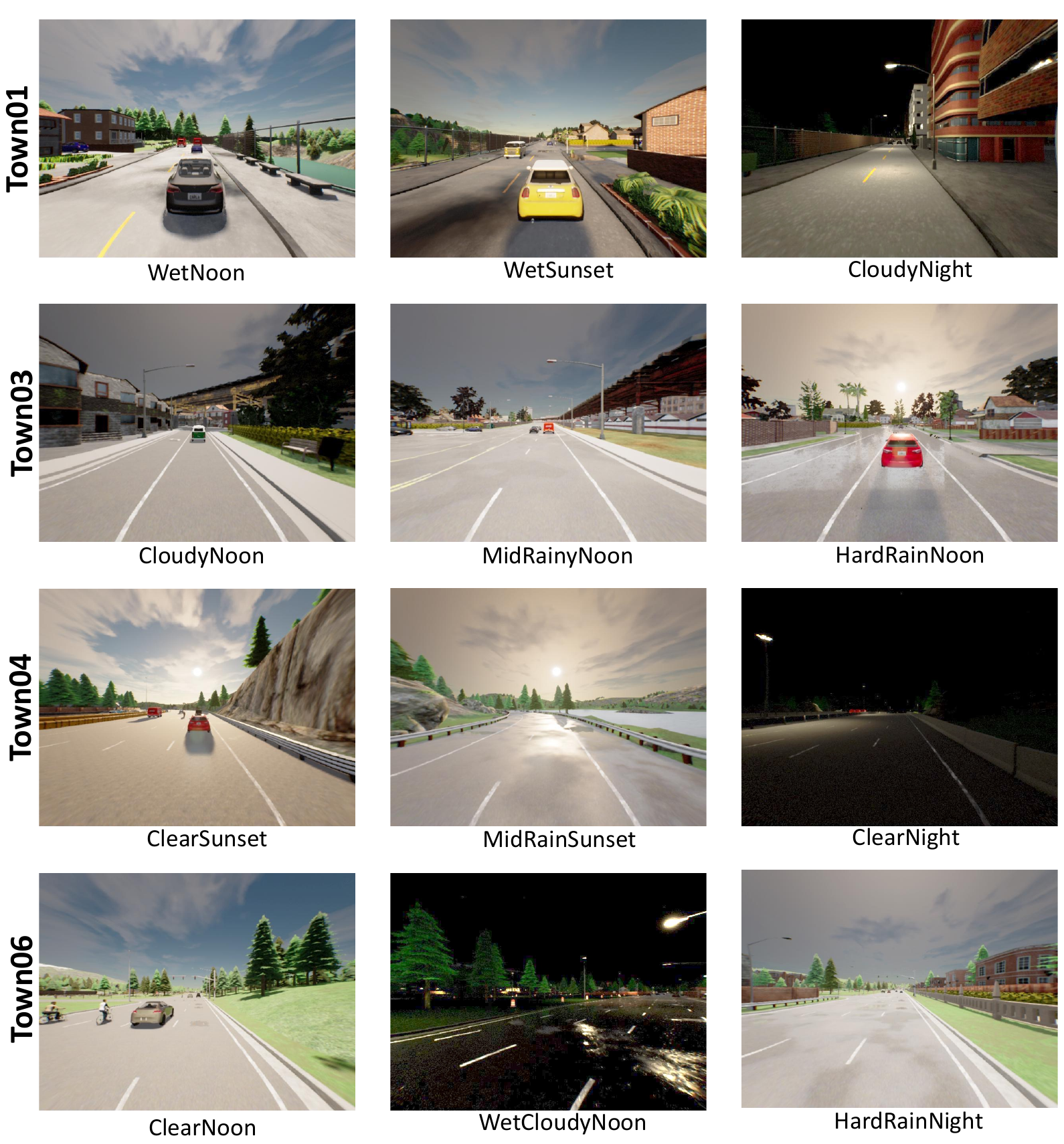}
  \caption{The visualization of some weather and daylight conditions used in the LangAuto Benchmark.}
  \label{fig:supp_vis_weathers}
\end{figure}

\begin{figure}
  \centering
  \includegraphics[width=\linewidth]{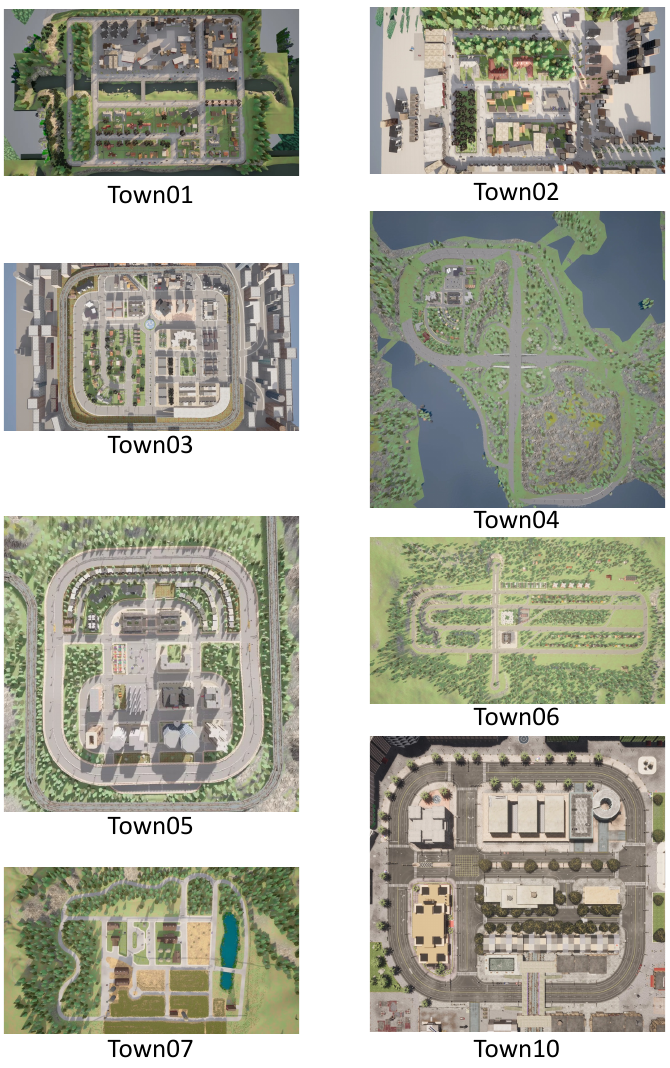}
  \caption{The visualization of eight town maps used in the LangAuto Benchmark.}
  \label{fig:supp_vis_town_map}
\end{figure}

\noindent \textbf{Model details} For the vision encoder, we use $K_{enc} = 1$ encoder layers and $K_{dec} = 3$ decoder layers. The feature of the 5-th stage in the ResNet is employed as the extracted feature map, then we apply an MLP layer to convert its dimension to 768, which is the feature dimension of the following Q-Former. Following LAV~\cite{chen2022learning}, we build a simplified version of PointNet~\cite{qi2018frustum} with several MLP layers and Batch Normalization layers to encode LiDAR point cloud data. For the Q-Former, we utilize the model architecture and pre-trained weights from the BLIP-2~\cite{li2023blip}. In our work, the visual tokens fed into the Q-Former include 400 BEV tokens, 5 future waypoint tokens and 1 traffic light token.

\vspace{1mm}
\noindent \textbf{Sensor configuration} We use one front-facing camera, two side-facing cameras, and one back-facing camera to collect RGB images. Each camera has a resolution of 800 $\times$ 600 and a $100^\circ$ horizontal field of view (FOV). The two side cameras are angled at $60^\circ$. For the font image, we scale the shorter side of the front camera input to 256 and crop its center patch of 224 $\times$ 224. For the focusing view image, we directly crop the center of the front camera input to get a 128 $\times$ 128 patch which can capture distance traffic light status. For the other images, the shorter side of the camera input is scaled to 160 and then takes a 128 $\times$ 129 center patch. For the LiDAR sensor, the rotation frequency is 10Hz and the upper/lower field-of-view is 10/-30. The number of channels is 64.

\vspace{1mm}
\noindent \textbf{Training details} We first introduce the details of the vision encoder pretraining. We adopt the AdamW optimizer~\cite{loshchilov2018decoupled} and a cosine learning rate scheduler~\cite{loshchilov2016sgdr} for the training. The initial learning rate set for the transformer encoder and 3D backbone is $\frac{Batch Size}{512} \times 5e^{-4}$. And $\frac{Batch Size}{512} \times 2e^{-4}$ is the learning rate for the 2D backbone because the backbone is initialized with the ImageNet pre-train weights. We train the models for 35 epochs with the first 5 epochs for warm-up~\cite{he2016deep}. We used random scaling from $0.9\  \text{to}\ 1.1$ and color jittering to augment the collected RGB images. 

Then we introduce the training details in the instruction-finetuning stage. We also adopt the cosine learning rate scheduler and the initial learning rate is $1e^{-4}$ for a batchsize 32. We train the models for 15 epochs with the first 2000 iteration steps for warm-up. The weight decay is set to 0.07. The maximum historic horizon $T_{max}$ is set to 40, and we will truncate the data clip to keep the recent 40 frames if its number of frames exceeds 40.  


\section{Additional Experiments}
In this section, we delve further into our method by conducting additional ablation studies. First, we assess our methodology using a variety of sample rates in Table~\ref{table:supp_sample_rate}. The term "sample rate" represents the fixed interval at which training frames are sampled.
When the sample rate is set at 1, the horizon becomes narrow, which prevents the application of temporal augmentation, and consequently leads to a decrease in performance. Conversely, setting the sample rate at 4 can create an excessively large gap between two consecutive frames, which might obtain a poor driving score. A sample rate of 2 achieves a good trade-off.

Second, we conduct ablation studies on the usage rate of notice instructions as shown in Table~\ref{table:supp_notice_rate}. In our method, we randomly removed 75\% notice instructions in the data clips to avoid overfitting. It's worth noting that we use the LangAuto-Short here, rather than  LangAuto-Notice, where the AV can not receive any notice instruction. We first remove all notice data, and get a worse driving score and infraction score. This suggests that incorporating notice instructions during training may improve the AV's ability to both attend to and understand adverse events, thus reducing collisions. However, when we tried to include all notice data, we found it did not enhance performance. Utilizing 25\% of the instructions achieves a good trade-off.

\begin{table}[]
\centering
\resizebox{0.9\linewidth}{!}{
\begin{tabular}{cccc}
\toprule
Sample rate  &  DS $\uparrow$  & RC $\uparrow$   &  IS $\uparrow$       \\ \cmidrule(r){1-1} \cmidrule(r){2-4} 
                  1       & 49.5$\pm$1.5   &  58.5$\pm$2.7  &  0.83$\pm$0.03  \\
                  2     & \textbf{50.6}$\pm$\textbf{1.7}   &  \textbf{60.0}$\pm$\textbf{3.4}    &  \textbf{0.84}$\pm$\textbf{0.04}      \\ 
                  4      & 46.0$\pm$2.1   &  59.5$\pm$2.7  & 0.79$\pm$0.03  \\ 
                        \bottomrule
\end{tabular}}

\caption{Ablation study on the different choices of the sample rate. The experiment is conducted on the LangAuto-Short benchmark with the backbone of LLaVA-v1.5.}

\label{table:supp_sample_rate}
\end{table}
\begin{table}[]
\centering

\resizebox{0.9\linewidth}{!}{
\begin{tabular}{cccc}
\toprule
Notice Data \%  &  DS $\uparrow$  & RC $\uparrow$   &  IS $\uparrow$       \\ \cmidrule(r){1-1} \cmidrule(r){2-4} 
                  0       & 45.2$\pm$2.8   &  \textbf{67.1}$\pm$\textbf{2.5}  &  0.68$\pm$0.03  \\
                  25     & \textbf{50.6}$\pm$\textbf{1.7}   &  60.0$\pm$3.4    &  \textbf{0.84}$\pm$\textbf{0.04}      \\ 
                  100      & 49.1$\pm$1.9   &  58.2$\pm$2.4  & 0.83$\pm$0.04  \\ 
                        \bottomrule
\end{tabular}}

\caption{Ablation study on the usage rate of notice instructions. The experiment is conducted on the LangAuto-Short benchmark with the backbone of LLaVA-v1.5.}

\label{table:supp_notice_rate}
\end{table}

\section{Benchmark Details}
In Figure~\ref{fig:supp_vis_weathers}, we show the 12 different environmental conditions (16 conditions are used in our benchmark in total). In Figure~\ref{fig:supp_vis_town_map}, we show all 8 town maps we used in this work. In Table~\ref{table:supp_benchmark_stats}, we list the basic statistical information of the LangAuto benchmarks across various tracks (LangAuto, LangAuto-Short, LangAuto-Tiny).

\begin{table}[h!]
\centering
\footnotesize
\resizebox{1\linewidth}{!}{
\begin{tabular}{l|ccc}
    \toprule
    Benchmark Type & LangAuto & LangAuto-Short & LangAuto-Tiny  \\
    \cmidrule(r){1-1}
    \cmidrule(r){2-4}
    Avg. Driving Distance (m)  & 635.8 & 305.9 & 122.4 \\
    \cmidrule(r){1-1}
    \cmidrule(r){2-4}
    Avg. Navigation Instructions & 20.3 & 10.8 & 5.1 	\\ 
    Avg. Notice Instructions & 5.8 & 3.3 & 1.7  \\
    \bottomrule
\end{tabular}}
\caption{Comparative Analysis of Different Benchmark Tracks}
\label{table:supp_benchmark_stats}
\end{table}

\section{Instruction Details}
Our work considers 56 different types of navigation and notice instructions, and we use ChatGPT to generate eight different phrases for the same navigation instruction. 
Table~\ref{table:supp_variants} shows an example where we generate eight phrases for the navigation instruction 'Turn Right'.
We present the full list of 56 different types of navigation and notice instructions in Table~\ref{table:supp_full_list}. Table~\ref{table:supp_misleading} demonstrates how we generate the misleading instructions. The first column is the driving scenario in which the agent is located, and the second column represents the possible misleading instructions generated by our framework. For example, when the AV is on a single-lane road, the misleading instruction ``\textit{Change your route to the left-hand lane}" violates the traffic rules and raises safety concerns. In Table~\ref{table:supp_connected}, we show some examples of the connected instructions.

\begin{table}[h]
\resizebox{0.9\linewidth}{!}{%
\begin{tabular}{l|l}
\toprule
\textbf{Instruction} & 
\textbf{Eight instructions of one kind of instruction}  \\
\cmidrule(lr){1-1}\cmidrule(lr){2-2}
\multirow{8}{*}{\textbf{Turn right}}
& After [x] meters, execute a right turn.  \\
& After [x] meters, take a right.  \\
& Right in [x] meters.  \\
& After [x] meters, hang a right.  \\
& In [x] meters, prepare to turn right.  \\
& Continue for [x] meters, then turn right.  \\
& Go [x] meters, then just take a right.  \\
& After [x] meters, a right turn is required.  \\
\cmidrule(lr){1-1}\cmidrule(lr){2-2}
\end{tabular}%
}
\caption{Eight different phrazes of one ``turn'' instruction generated by the ChatGPT API.}
\label{table:supp_variants}
\end{table}

\begin{table*}[h]
\centering
\resizebox{0.9\linewidth}{!}{%
\begin{tabular}{c|l}
\toprule
\hspace{1cm} \textbf{Type}  \hspace{1cm} & 
\textbf{One randomly chosen instruction of each instruction type}  \\
\cmidrule(lr){1-1}\cmidrule(lr){2-2}
\multirow{16}{*}{\textbf{Follow}}
& Transition to the left lane for travel.  \\
& You might want to switch to the right lane.  \\
& In about [x] meters, you'll want to switch to the left lane.  \\
& In [x] meters, reposition to the right lane.  \\
& Keep going on this road, you're doing great!  \\
& Continue on the highway.  \\
& Maintain your course along this route for precisely [x] meters.  \\
& Cruise down the highway for about [x] meters.  \\
& Continue in a straight line along your current path.  \\
& Keep going straight until you reach the next junction, you're on the right track!  \\
& Preserve your current trajectory for exactly [x] meters.  \\
& Stay straight for [x] meters until the next intersection.  \\
& Veering to the left, prepare to enter the highway.  \\
& Execute a right maneuver, prepare for highway exit.  \\
& In [x] meters, slide left and plan to hop off the highway.  \\
& In [x] meters, proceed to the right, prepare for immediate highway departure.  \\
\cmidrule(lr){1-1}\cmidrule(lr){2-2}
\multirow{25}{*}{\textbf{Turn}}
& Prepare to turn left up ahead.  \\
& Proceed ahead and make a right turn.  \\
& Continue for [x] meters, then turn left.  \\
& After [x] meters, execute a right turn.  \\
& You're going to be turning left at the next junction, alright?  \\
& It is mandatory to take a right turn at the imminent intersection.  \\
& Straight through the next crossroads.  \\
& After navigating [x] meters, a left turn at the intersection is obligatory.  \\
& After moving forward [x] meters, prepare to make a right turn at the intersection.  \\
& [x] meters more, then straight on at the intersection, piece of cake.  \\
& When you reach the next traffic signal, you will need to turn left.  \\
& Please execute a right turn upon reaching the upcoming traffic signal.  \\
& Please maintain your course straight at the next traffic signal.  \\
& Just another [x] meters, then you'll be turning left at the light, okay?  \\
& After [x] meters, take a right at the light.  \\
& After traversing [x] meters, it's crucial to continue straight at the traffic signal.  \\
& Next T-junction, turn left.  \\
& At the forthcoming T-intersection, execute a right turn.  \\
& Just keep on going straight through the next T-junction, sound good?  \\
& In [x] meters, hang a left at the T.  \\
& After [x] meters, take a right at the T, no biggie.  \\
& After a distance of [x] meters, maintaining a straight course at the T-intersection is imperative.  \\
& Find your way out at the first exit on the roundabout, please.  \\
& Depart at the second exit on the roundabout.  \\
& Shoot out on the third exit.  \\
\cmidrule(lr){1-1}\cmidrule(lr){2-2}
\multirow{5}{*}{\textbf{Others}}
& Feel free to start driving.  \\
& Please implement an immediate reduction in your speed.  \\
& Hit the brakes, stop now.  \\
& Drive freely.  \\
& Please navigate towards the designated point, which is [x] meters in front of you and [y] meters to your left/right.  \\
\cmidrule(lr){1-1}\cmidrule(lr){2-2}
\multirow{10}{*}{\textbf{Notice}}
& Please watch out for the pedestrians up ahead.  \\
& Attention is required for the bicycle ahead.  \\
& Watch for the car that's just stopped up front.  \\
& Be mindful of the vehicle crossing on a red light to your left.  \\
& Car ran red light ahead.  \\
& Please be alert of the uneven road surface in the vicinity ahead.  \\
& Watch for the tunnel coming up.  \\
& Just a heads up, there's a red light ahead.  \\
& Green light ahead.  \\
& Be mindful of the yellow light ahead.  \\
\cmidrule(lr){1-1}\cmidrule(lr){2-2}
\end{tabular}%
}
\caption{Full list of the 56 different types of navigation instructions and notice instructions considered in our framework.}
\label{table:supp_full_list}
\end{table*}

\begin{table*}[h]
\centering
\resizebox{0.8\linewidth}{!}{%
\begin{tabular}{c|l}
\toprule
\textbf{Driving Scenarios} & 
\textbf{One randomly chosen instruction of each misleading type}  \\
\cmidrule(lr){1-1}\cmidrule(lr){2-2}
\multirow{8}{*}{\textbf{Single-lane Road}}
& Change your route to the left-hand lane.  \\
& Transition to the right lane for travel.  \\
& Proceed ahead and make a left turn.  \\
& A right turn is required up ahead.  \\
& Hang a left at the next crossroads.  \\
& Depart at the first exit on the roundabout.  \\
& Roll out on the second exit.  \\
& It is necessary for you to take the third exit on the roundabout.  \\
\cmidrule(lr){1-1}\cmidrule(lr){2-2}
\multirow{3}{*}{\textbf{Non-highway Road}}
& Maintain your course along the highway.  \\
& Slide left and plan to hop on the highway.  \\
& Ease on to the right and get set to quit the highway.  \\
\cmidrule(lr){1-1}\cmidrule(lr){2-2}
\multirow{3}{*}{\textbf{Non-roundabout Road}}
& First exit.  \\
& No sweat, just hit the second exit on the roundabout.  \\
& It is necessary for you to take the third exit on the roundabout.  \\
\cmidrule(lr){1-1}\cmidrule(lr){2-2}
\multirow{7}{*}{\textbf{Non-intersection Road}}
& Just up ahead, take a left. \\
& Your next action is a right turn, just ahead. \\
& Execute a left maneuver, prepare for highway entry. \\
& Right, ready to exit.  \\
& Carefully navigate to the first exit as you approach the roundabout.  \\
& You are required to take the second exit on the roundabout.  \\
& Third exit.  \\
\cmidrule(lr){1-1}\cmidrule(lr){2-2}
\multirow{3}{*}{\textbf{T-Intersection (Left Turn Prohibited)}}
& You'll be turning left at the next T-junction, alright?  \\
& Transition to the left lane for travel.  \\
& You might want to switch to the right lane.  \\
\cmidrule(lr){1-1}\cmidrule(lr){2-2}
\multirow{3}{*}{\textbf{T-Intersection (Right Turn Prohibited)}}
& A right turn is mandatory at the upcoming T-intersection.  \\
& Change your route to the left-hand lane.  \\
& Just head for the right lane.  \\
\cmidrule(lr){1-1}\cmidrule(lr){2-2}
\multirow{2}{*}{\textbf{When turning}}
& Please adjust your course to the left-most lane.  \\
& Reposition to the right lane.  \\
\bottomrule
\end{tabular}%
}
\caption{Full list of the misleading instructions and their corresponding driving scenarios.}
\label{table:supp_misleading}
\end{table*}

\begin{table*}[h]
\centering
\resizebox{0.8\linewidth}{!}{%
\begin{tabular}{c|l}
\toprule
\textbf{Driving Scenarios} &
\textbf{Examples of connected instructions}  \\
\cmidrule(lr){1-1}\cmidrule(lr){2-2}
\multirow{2}{*}{\textbf{Two consecutive instructions}}
& (1) Prepare to turn right up ahead. (2) Proceed along this route.  \\ \cmidrule(lr){2-2}
& (1) Maintain your course along this route. (2) Left, ready to enter. \\
\cmidrule(lr){1-1}\cmidrule(lr){2-2}
\multirow{6}{*}{\textbf{Three consecutive instructions}}
& (1) It's critical to keep straight at the forthcoming T-intersection.\\
& (2) Keep cruising down this road. (3) At the next traffic signal, you should make a right turn.  \\
\cmidrule(lr){2-2}
& (1) At the forthcoming T-intersection, execute a right turn.\\
& (2) Just head for the left lane.(3) Maintain your course along this route.  \\ \cmidrule(lr){2-2}
& (1) Keep going on this road, you're doing great! (2) Right ahead. (3) Right, ready to exit.  \\
\bottomrule
\end{tabular}%
}
\caption{Examples of the connected navigation instructions considered in the LangAuto benchmark.}
\label{table:supp_connected}
\end{table*}

\end{document}